# MULTIPLE OBJECTS TRACKING IN SURVEILLANCE VIDEO USING COLOR AND HU MOMENTS

Chandrajit M[1], Girisha R[2] and Vasudev T[1]

[1]Maharaja Research Foundation, Maharaja Institute of Technology, Mysore, India
[2]PET Research Foundation, PES College of Engineering, Mandya, India

## ABSTRACT

*Multiple objects tracking finds its applications in many high level vision analysis like object behaviour interpretation and gait recognition. In this paper, a feature based method to track the multiple moving objects in surveillance video sequence is proposed. Object tracking is done by extracting the color and Hu moments features from the motion segmented object blob and establishing the association of objects in the successive frames of the video sequence based on Chi-Square dissimilarity measure and nearest neighbor classifier. The benchmark IEEE PETS and IEEE Change Detection datasets has been used to show the robustness of the proposed method. The proposed method is assessed quantitatively using the precision and recall accuracy metrics. Further, comparative evaluation with related works has been carried out to exhibit the efficacy of the proposed method.*

## 1. INTRODUCTION

Visual surveillance is one of the most active research areas in computer vision. An automatic visual surveillance system generally contains computer vision tasks such as motion segmentation, object classification, tracking, recognition and high-level motion analysis [1, 2]. Motion segmentation task detects and extracts moving objects from video frames. Object classification task categorizes the detected objects into predefined classes such as human, vehicle, animal, clutter, etc. Object tracking task establishes the state of object in the successive frames of a video. Further, object tracking can be performed on a single object or multiple objects using intra-camera or inter-camera setup. The intra-camera approach uses one camera to track the objects within its Field Of View (FOV) and inter-camera approach uses multiple cameras to track objects in overlapping FOV [3]. Finally, the high-level motion analysis task recognizes the objects and interprets their activities.

Object tracking has a wide variety of applications like smart video surveillance, traffic video monitoring, accident prediction and detection, motion-based recognition, human computer interaction, human behavior understanding, etc. In addition, the constant threats from terrorist attacks at public and secured places increase the need for efficient surveillance systems with embedded object tracking subsystems. Further, the advancement in computing technology and availability of high quality cameras at low cost has increased the use of object tracking in numerous applications [1-4].

                                                                    15

Object tracking in general, is a very challenging task because of the following circumstances; complex object shapes, cluttered background, loss of information caused by representing the real-world 3D into a 2D scene, noise in images induced by image capturing device, illumination variations, occlusions, shadows, etc. [1, 5, 6].

This paper presents a feature based method for tracking multiple moving objects in video sequences captured in complex environment. The paper is organized as follows: Section 2 reports the related work on object tracking, Section 3 presents the overview of proposed work, Section 4 presents the proposed object tracking algorithm followed by experimental results and conclusion in Section 5 and 6 respectively.

## 2. LITERATURE REVIEW

Multiple objects tracking involve the association of segmented motion objects in the successive frames of a video sequence. Generally, this association is done by matching the features of segmented foreground objects. Features such as centroid, motion field, stereo disparity, edges, texture, colors or gradients are commonly extracted from the motion object blob for association. Further, while tracking in cluttered environment the motion object blob may be occluded and hence the features cannot be extracted from the occluded motion object blob. In such a case, prediction tools such as Kalman filter or particle filter are used to estimate the object location [1, 5]. Kalman filter is employed when the data associated to the object is assumed to be Gaussian distributed. Kalman filter uses three steps namely initialization, prediction and update to predict the state of object in Bayesian framework. The particle filter is a variant of Kalman filter which is used for non-Gaussian object state and usually employed for tracking multiple objects.

Because of the challenges involved in tracking objects as discussed in previous section, a lot of work have been carried out on object tracking and these approaches can be classified as region based [7-9] , feature based [10-13] , model based [14-20] and hybrid [21-25]. The region based methods track object by observing the regions of motion object. The feature based methods track object using the features of motion object. The model based methods track the object by constructing an apriori model of the motion object. The hybrid based methods combines the above mentioned methods to track object in frames of a video [1-5].

Region based tracking across three cameras using Kalman filter is proposed in [7]. Girisha and Murali [8, 9] adopted optical flow based method for object tracking using two-way ANOVA to compare extracted features of video frames. However, the algorithm does not maintain the identity of the tracked objects. A feature based tracking method in which corner points of extracted vehicles are used for tracking is proposed in [10]. A feature based head tracker that uses color information and particle filter is proposed in [11] for single object tracking. Another feature based method that uses the edges of object is proposed in [12]. However, the method is designed only for tracking the face of a person. Wang et al. [13] propose a scene adaptive feature based method to track multiple objects.

Martin and Martinez [14] used color based appearance, motion information and adaptive particle filter to track object. Kristan et al. [15] computed local motion of target and used color based particle filter for tracking objects. However, the model cannot be used for all practical applications as the object has to be selected manually and is computationally expensive. A learning based method that learns color, size and motion to track objects across cameras using Kalman filter is proposed in [16]. A well-known work namely Hydra [17] tracks head by template matching approach. A model based method that uses Principle axis of an object and Homography constraints to match and track the object across different views of camera in Kalman filter framework is proposed in [18]. However, the method may not be applicable for practical applications as landmarks on the ground plane are required and the system has the limitation to work in controlled environment. Xi et al. [19] propose a multiple objects tracking algorithm in which object association in successive frames is based on integer programming flow model. However, the algorithm cannot be used for real-time scenario as the accuracy decreases in cluttered environment. Jahandide et al. [20] propose an algorithm to track single moving object. They tracked the object using appearance and motion model using Kalman filter. However, the user needs to provide the objects' position in the first frame for tracking.

A hybrid approach is proposed in [21] where statistical method is combined with CRF (Conditional Random Field) framework and sliding window optimization algorithm for labeling objects. Another hybrid approach that combines SSD (Square Sum of Difference) and color based mean-shift tracker in Kalman filter framework is proposed in [22]. Thome et al. [23] combined region based with appearance based methods to track motion object. Ali and Dailey [24] tracked multiple human in crowd using Adaboost classifier and particle filter. Face detection based on Haar-Like features (machine learning based clustering technique) and mean-shift tracking is proposed in [25].

The region based tracking methods cannot reliably handle occlusions. Further, the region based methods do not accurately track in the case of a cluttered environment with multiple moving objects. The feature based methods overcome the occlusion limitation of region based methods. However, the efficiency of feature based method decreases in the case of distortion during perspective projection. The model based methods have some shortcoming such as the necessity of constructing models and high computational cost. The hybrid based methods are suitable for tracking objects in a complex environment. However, a suitable combination of methods to make a hybrid model is a challenging task.

Although there are several methods proposed in the literature for objects tracking. These methods are based on constraints and assumptions such as; requires a training step, works for constrained environment, designed to track only part of an object like face, hand etc. and assumes no occlusion scenario. These constraints and assumptions may not be suitable for real-time environment. The proposed work aims to overcome some of the limitations addressed above by devising an algorithm to track multiple moving objects in cluttered environment which tracks objects by establishing the identity in the successive frames of a video sequence.

## 3. OVERVIEW OF THE PROPOSED METHOD

The proposed multiple objects tracking block diagram is shown in Fig.1 as an iterative sequence of the motion segmentation and tracking phase. In the motion segmentation phase the motion objects are segmented using the method proposed in [26]. Subsequently, a hole filling and noise elimination process is applied on the motion segmented frames for cleaning up the noise contained in the object blobs. Following the above process, the color and Hu moments, and area features are extracted from the segmented motion object blob using connected component analysis. The color and Hu moments are used for object representation and the area of object blob is used for elimination of smaller object (noise). Finally, in the tracking phase the extracted features of the individual objects in the successive frames are matched using Chi-Square dissimilarity measure with nearest neighbor classifier to track the moving objects. In addition, each tracked object within the FOV of a static camera will be given an identity. Further, a new object entering the FOV or an existing object leaving the FOV of a camera is also addressed and the identity is updated.

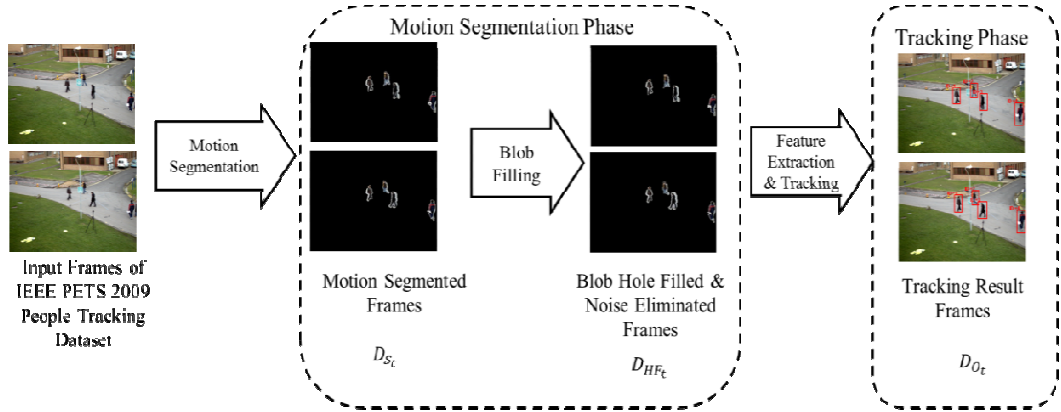

Figure 1. The block diagram of the proposed method.

# 4. THE PROPOSED METHOD

## 4.1 Motion Segmentation using Chi Square Statistics

Motion segmentation is the initial task that is carried out before the object tracking stage. In this paper, motion segmentation is performed in temporal domain using the method [26] in which a pixel is decided as a motion pixel by employing the Chi-Square statistical test on the $3 \times 3$ pixel neighborhood. Subsequently, the motion segmented frames $D_{S_t}$ and $D_{S_{t+1}}$ are generated, where, $t$ is current temporal frame. A blob hole filling process is applied to eliminate any holes in the moving blob using the method proposed in [27]. In the next step, a morphological erosion operation is applied to eliminate the remaining noise. Finally, the blob hole filled and noise suppressed frames $D_{HF_t}$ and $D_{HF_{t+1}}$ are generated.

## 4.2 Feature extraction

Feature representation of the object plays a key role in tracking the state of the object in the frames of a video. The commonly used features to represent the object are area, centroid, major axis, minor axis, corners, textures etc. Moreover, the choice of features varies according to the tracking application. In order to track an object which is very small, the centroid feature is usually appropriate. On the other hand, to track an object which is big in size, the features such as motion field, stereo disparity, edges, texture, colors, contours, gradients etc., can be used to represent the object. In this work, color and Hu moments and area features are used for object representation. The features are extracted from the segmented motion object blob of the frames $D_{HF_t}$ and $D_{HF_{t+1}}$ after applying a connected component analysis as shown in Fig 2. The extracted features are recorded in the feature vectors. Table. 1 and Table. 2 show the example of extracted features from the segmented objects in the frame shown in Fig. 2.

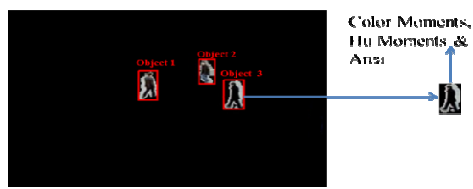

Figure 2. Feature extraction

Table 1. Area and Color Moments Feature Values.

| Object | Area | Color Moments | | | | | | | | |
|---|---|---|---|---|---|---|---|---|---|---|
| | | $\mu_R$ | $\sigma_R$ | $s_R$ | $\mu_G$ | $\sigma_G$ | $s_G$ | $\mu_B$ | $\sigma_B$ | $s_B$ |
| 1 | 592 | 65.79 | 72.98 | 0.69 | 64.12 | 73.24 | 0.73 | 63.17 | 72.46 | 0.75 |
| 2 | 433 | 57.71 | 62.40 | 0.62 | 58.26 | 63.70 | 0.60 | 58.58 | 64.02 | 0.57 |
| 3 | 434 | 67.81 | 78.60 | 0.53 | 68.46 | 79.68 | 0.54 | 68.62 | 79.68 | 0.53 |

Table 2. Hu Moments Feature Values.

| Object | Hu Moments | | | | | | |
|---|---|---|---|---|---|---|---|
| | $\phi_1$ | $\phi_2$ | $\phi_3$ | $\phi_4$ | $\phi_5$ | $\phi_6$ | $\phi_7$ |
| 1 | 0.213 | 0.014 | 0.001 | 3.793 | 2.472 | 4.229 | 3.578 |
| 2 | 0.202 | 0.011 | 0.000 | 2.156 | 1.821 | 1.168 | -1.298 |
| 3 | 0.231 | 0.022 | 0.000 | 2.842 | -1.209 | -4.005 | 2.017 |

The color moments aid in representing the object and also used in image retrieval applications. Furthermore, the color moments are scaling and rotation invariant and hence, they can be used for varying illumination video sequences. Generally, the lower order color moments namely mean, standard deviation and skewness are used as features for applications [28]. In this work, we are using the same features to represent the object blob. The first three color moments are computed for the RGB color channels of each pixel $P$ by using the following equations respectively. Thereby, nine color feature values are extracted from the segmented motion object blob.

$$\mu_c = \sum_{i=1}^{N} \frac{1}{N} P(w,h)_c \qquad (1)$$

where, C={R,G,B} and N=number of pixels in object blob

$$\sigma_c = \sqrt{\frac{1}{N}\sum_{i=1}^{n}(P(w,h)_c - \mu_c)^2} \qquad (2)$$

$$s_c = \sqrt[3]{\frac{1}{N}\sum_{i=1}^{N}(P(w,h)_c - \mu_c)^3} \qquad (3)$$

It has been observed that using the color moments alone cannot achieve better tracking results as shown in Fig. 3a. In this work, an ensemble of color and Hu moments are used to represent the object blob which gives significantly better tracking results as shown in Fig. 3b.

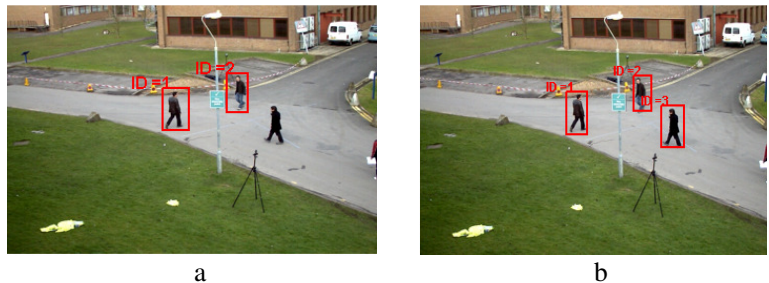

a        b

Figure 3. a) Tracking results using color moments b) Tracking results using color and Hu moments

Image moments have been extensively used in image analysis because of its robustness to translation, scaling and rotation [29,30]. In order to represent the object blob by good features which will help in robust tracking, Hu moments of object blob are also extracted. Based on the normalized central moments the seven Hu moments invariant for object blob is computed using the equations (4) through (10).

$$\phi_1 = (\eta_{20} + \eta_{02}) \qquad (4)$$

$$\phi_2 = (\eta_{20} - \eta_{02})^2 + 4\eta_{11}^2 \qquad \text{`} \qquad (5)$$

$$\phi_3 = (\eta_{30} - 3\eta_{12})^2 + (3\eta_{21} - \mu_{03})^2 \qquad (6)$$

$$\phi_4 = (\eta_{30} + \eta_{12})^2 + (\eta_{21} + \mu_{03})^2 \qquad (7)$$

$$\phi_5 = (\eta_{30} - 3\eta_{12})(\eta_{30} + \eta_{12})[(\eta_{30} + \eta_{12})^2 - 3(\eta_{21} + \eta_{03})^2] + (3\eta_{21} - \eta_{03})(\eta_{21} + \eta_{03})[3(\eta_{30} + \eta_{12})^2 - (\eta_{21} + \eta_{03})^2] \qquad (8)$$

$$\phi_6 = (\eta_{20} - \eta_{02})[(\eta_{30} + \eta_{12})^2 - (\eta_{21} + \eta_{03})^2] + 4\eta_{11}(\eta_{30} + \eta_{12})(\eta_{21} + \eta_{03}) \qquad (9)$$

$$\phi_7 = (3\eta_{21} - \eta_{03})(\eta_{30} + \eta_{12})[(\eta_{30} + \eta_{12})^2 - 3(\eta_{21} + \eta_{03})^2] - (\eta_{30} - 3\eta_{12})(\eta_{21} + \eta_{03})[3(\eta_{30} + \eta_{12})^2 - (\eta_{21} + \eta_{03})^2] \qquad (10)$$

Where, $\eta_{pq} = \frac{\mu_{pq}}{\mu_{00}^{\gamma}}, \gamma = \frac{(p+q+2)}{2}, p+q = 2,3, \dots$ and $\mu_{pq}$ are the invariant features of centralized moments which are defined as in equation (11).

$$\mu_{pq} = \int_{-\infty}^{+\infty} \int_{-\infty}^{+\infty} (x - \bar{x})^p (y - \bar{y})^q f(x,y) \, dx \, dy \qquad (11)$$

where, $p, q = 0,1,2 \dots, \bar{x} = \frac{m_{10}}{m_{00}}, \bar{y} = \frac{m_{01}}{m_{00}}$ and the pixel point $(\bar{x}, \bar{y})$ are centroid of the object blob.

After the motion segmentation stage, there may by some noise whose area will be considerably small is size. Hence, to eliminate such unwanted objects the area feature is also extracted from the motion object blob.

### 4.3 Multiple Objects tracking

Tracking of multiple objects means associating each of the segmented objects in the previous frame with the corresponding object in the next consecutive frame of a video sequence. The association is achieved by correlating the features of moving objects in the successive frames. In this work, Chi-Square dissimilarity measure [31] is used to measure the degree of association of the objects features in the successive frames. The feature vectors which consists of color and Hu moments features of segmented moving objects in two frames are used to find the similarity of objects. An object in frame $D_{HF_t}$ is assigned to an object in frame $D_{HF_{t+1}}$ that minimizes the Chi-Square distance. A nearest neighbor classifier with Chi-Squared distance is used to map the object identity as shown in Eq. 12. Subsequently, for each object within the FOV of a camera the identity (numbered) will be established. Moreover, the objects in the FOV of a camera may enter or exit from any direction. In such a case, the new object entering the FOV will be given a new identity and the identity of the object which exits the FOV will be eliminated.

$$D(o_t, o_{t+1}) = \sum_{n=1}^{N} \frac{(o_{t_n} - o_{t+1_n})^2}{o_{t_n} + o_{t+1_n}} \qquad (12)$$

where, $N$=16 represent the nine color and seven Hu moments and $o_{t_n}$ and $o_{t+1_n}$ are respectively the feature values of the object in the frames $D_{HF_t}$ and $D_{HF_{t+1}}$ respectively at the $n^{th}$ feature.

## 5. EXPERIMENTS

The algorithm is tested on more than 70,000 frames of the benchmark IEEE PETS [32] and IEEE CHANGE DETECTION (CD) 2014 [33] dataset. The dataset includes both the indoor and outdoor environment sequences with a varied number of objects. The example of results for the sequences of IEEE PETS 2009, 2006 and sequences of IEEE CD Bungalows, Backdoor, Copy Machine dataset are shown in Fig. 4-9 respectively. Where, the top row is the output of motion segmentation phase and the bottom row is the output object tracking phase. In the output frame, the tracked objects are numbered according to the identity once they enter the FOV of a camera and surrounded by a red bounding box.

In the output frames Fig. 6 (top row) of motion segmentation, some motion objects segmented with very minimum area (shown using red arrow) are not tracked. This is resulted due to the constraint set on the area for the object in the proposed method as discussed in earlier section. However, even the objects whose area is less than the constraint set can be tracked.

The output of motion segmentation stage plays a major role in accuracy of tracking results. Fig.10 shows the output obtained for tracking by using the resultant frames of motion segmentation algorithm and manual segmentation. The result corroborates that the accuracy of the proposed is better in the case of accurate motion segmentation. Further, the presence of shadow in the motion object blob hinders the tracking results. Fig. 11 is an example in which the shadow of the object is also given an identity and tracked. Hence, the incorporation of shadow segmentation task will improve the tracking results of the proposed method.

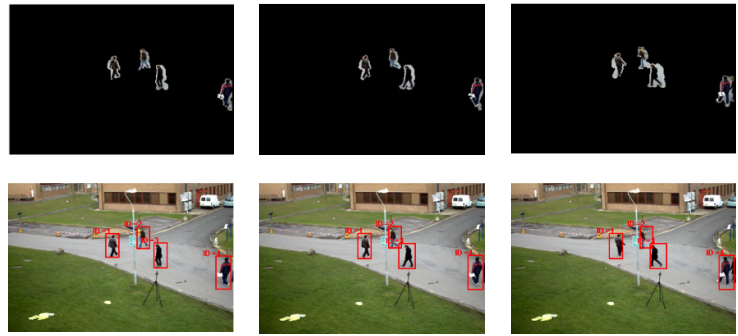

Figure 4. Tracking results of IEEE PETS 2009 S2.L1 View_001 dataset for frames frame00013-15

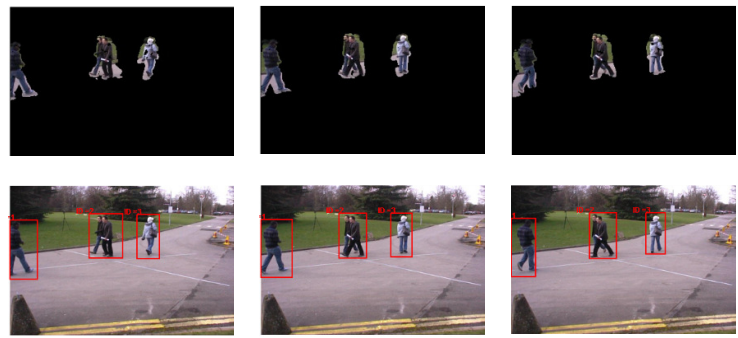

Figure 5. Tracking results of IEEE PETS 2009 S2.L1 View_007 dataset for frames frame000148-150

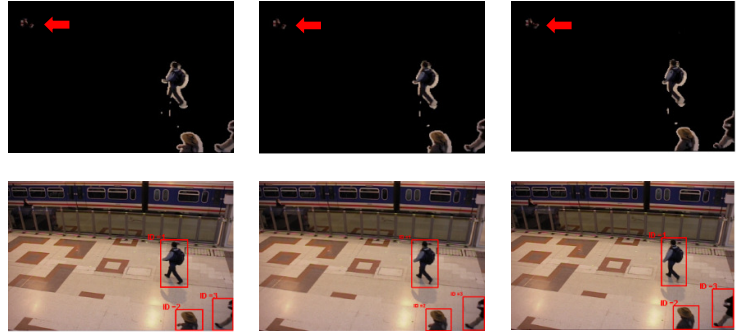

Figure 6. Tracking results of IEEE PETS 2006 dataset for frames in000052-54

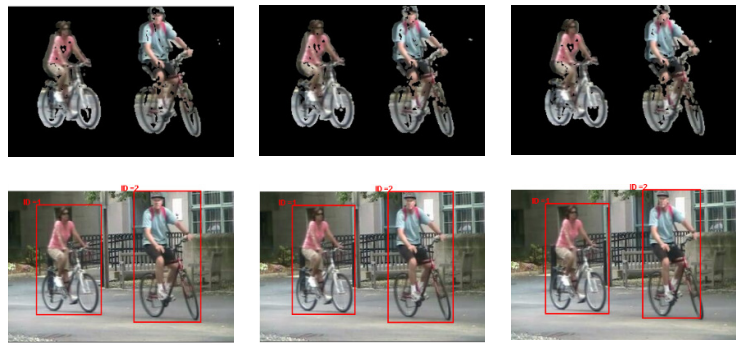

Figure 7. Tracking results of IEEE CD Backdoor dataset for frames in000100-102

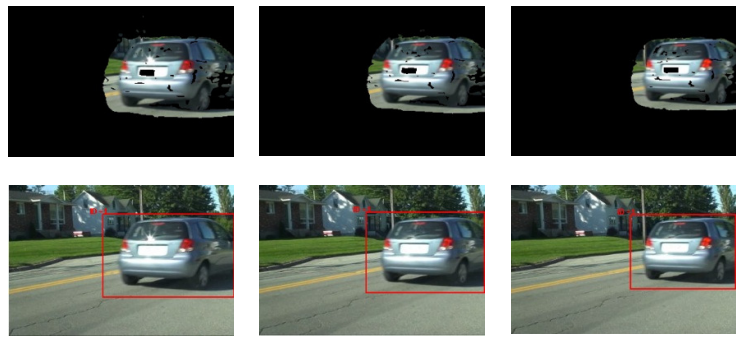

Figure 8. Tracking results of IEEE CD Bungalows dataset for frames in000320-322

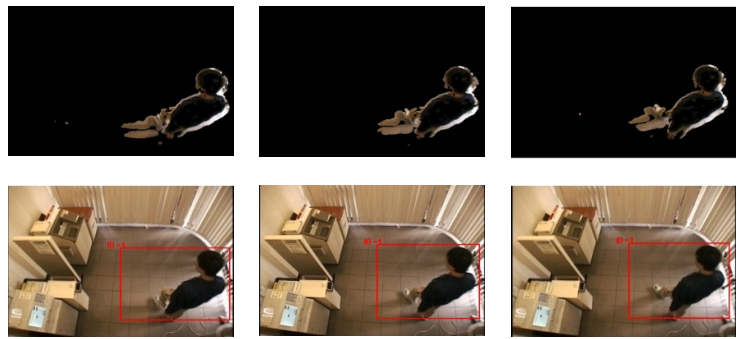

Figure 9. Tracking results of IEEE CD CopyMachine dataset for frames in000088-90

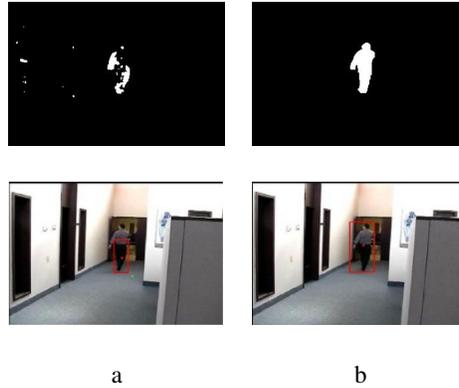

a                                          b

Figure 10. Tracking results of CD Cubicle dataset for frame in001268 (a) Result for motion segmentation using [26] (b) Result for manually segmented foreground

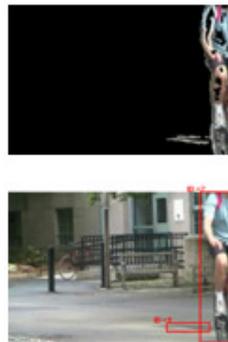

Figure 11. Example of tracking in presence of shadow (CD Backdoor dataset)

To corroborate the efficacy of the proposed method, performance has been assessed by the precision and recall accuracy metrics [1]. The precision and recall can be defined as follows:

$$Precision = \frac{No.\ correct\ correspondences}{No.\ established\ correspondences},$$ (13)

$$Recall = \frac{No.\ correct\ correspondences}{No.\ actual\ correspondences},$$ (14)

where, the actual correspondence denotes the correspondence as in the ground truth.

The results in Table 3. shows that the algorithm is successful in tracking the multiple objects with average precision and recall of 0.8483 and 0.8011 respectively. The sequences IEEE CD Copy Machine contain ghosts which are resulted because of non-stationary objects within the FOV of a camera. These ghost objects are considered as background in the motion segmentation process and therefore not segmented. Hence, recall is comparatively less for such dataset.

Comparative assessment with the state-of-the-art methods has been done and shown in the Table 4. Furthermore, Fig. 12 shows the comparison of the results with Kalman filter based method available in MATLAB R2011a (People Tracking Demo). The results validate that the proposed method performs better compared to the Kalman filter based method.

23

Table 3. Performance accuracy of proposed method

| Dataset | Scene | No. objects | Precision | Recall |
|---|---|---|---|---|
| IEEE PETS 2006 | Indoor | 15 | 0.8707 | 0.7966 |
| IEEE PETS 2009 | Outdoor | 37 | **0.9418** | 0.7168 |
| IEEE CD Bungalows | Outdoor | 09 | 0.9127 | 0.8543 |
| IEEE CD Backdoor | Outdoor | 05 | 0.8666 | **1.0000** |
| IEEE CD Copy Machine | Indoor | 05 | **0.6470** | **0.6470** |
| IEEE CD Cubicle | Indoor | 02 | 0.8510 | 0.7920 |
| Average | | | **0.8483** | **0.8011** |

Table 4. Empirical comparison of tracking methods

| Authors | Approach | Entry | Exit | No. of Objects | Occlusion | Identity |
|---|---|---|---|---|---|---|
| Girisha and Murali[9] | Region Based | No | No | Multiple | Yes | No |
| Nummiaro et al.[11] | Feature Based | - | - | Single | Yes | No |
| Martin and Martinez[13] | Model Based | - | - | Multiple | Yes | Yes |
| Heili et al.[18] | Hybrid | Yes | Yes | Multiple | No | Yes |
| Proposed | Feature Based | Yes | Yes | Multiple | No | Yes |

The time required for the processing a single frame of resolution 360 x 240 by each stage of the proposed algorithm is shown in Table 5. This test was carried using Intel Dual Core 1.8 GHz machine with Windows 7 Operating System and implemented using MATLAB R2011a.

Table 5. Computational time

| Stage | Time in secs |
|---|---|
| Motion Segmentation (two frames) | 0.12 |
| Blob hole filling and Morphological processing | 0.14 |
| Tracking | 0.42 |
| Total | 0.68 |

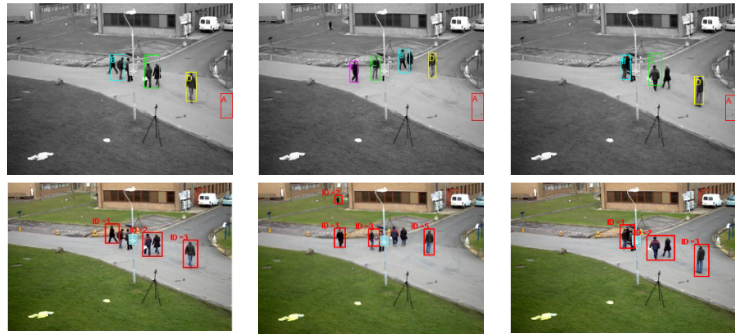

Figure 12**.** Comparative Tracking Results for Frames of IEEE PETS 2009 S2L1 Dataset (Top Row) Kalman Filter Result (Bottom Row) Proposed Method Result.

## 6. CONCLUSION

In this paper, a new method to track the multiple moving objects by ensemble of color and Hu moments is proposed. A nearest-neighbor classifier with Chi-Square dissimilarity measure is applied to track the objects in successive frames. Further, each moving object within the FOV of a camera is given an identity thereby establishing the recognition of the object in the successive frames of a video.

The method had been tested using the benchmark IEEE PETS and IEEE Change Detection dataset to prove its efficacy. Quantitative evaluation has been done with precision and recall tracking accuracy metrics. Furthermore, comparative assessment with state-of- the-art methods is done to show the superiority of the proposed method. The proposed method is able to perform better compared to the Kalman filter based method. In future, occlusion handling strategy and shadow elimination will be considered for the improved tracking results.

## AUTHORS

**Chandrajit M,** received his MCA degree from University of Mysore in 2007.He is pursuing Ph.D in Computer Science and Technology from University of Mysore, India. He is having 10 years of experience in teaching and industry. His research interests include computer vision, digital image processing and pattern recognition.

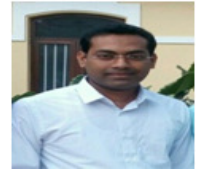

**Girisha R,** is Professor in the Department of Computer Science and Engineering, PES College of Engineering, Mandya, India. He obtained his Master's Degree in Computer Science and Technology from University of Mysore. He was awarded Ph.D in Computer Science from University of Mysore. He is having 15 years of experience in teaching and research. His research interests include computer vision, digital image processing and pattern recognition.

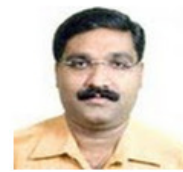

**Vasudev T,** is Professor in the Department of Computer Applications, Maharaja Institute of Technology, Mysore, India. He obtained his Bachelor of Science and Post Graduate diploma in Computer Programming and also two Master's degree one in Computer Applications and the other one in Computer Science and Technology. He was awarded Ph.D. in Computer Science from University of Mysore. He is having 30 years of experience in academics. His research interests include digital image processing specifically document image processing, computer vision and pattern recognition.

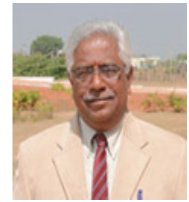